\documentclass[11pt]{article}

\usepackage[final]{acl}

\usepackage{times}
\usepackage{latexsym}

\usepackage[T1]{fontenc}

\usepackage[utf8]{inputenc}
\usepackage{multirow}
\usepackage{amsmath}
\usepackage{microtype}
\usepackage{multirow}
\usepackage{enumitem}

\usepackage{inconsolata}

\usepackage{graphicx}

\usepackage{dblfloatfix}

\title{SwanNLP at SemEval-2026 Task 5: An LLM-based Framework for Plausibility Scoring in Narrative Word Sense Disambiguation}

\author{
    Deshan Sumanathilaka, Nicholas Micallef, Julian Hough, Saman Jayasinghe \\
    Department of Computer Science, Swansea University, Wales, UK \\
    { \{t.g.d.sumanathilaka, nicholas.micallef, julian.hough, s.j.j.g.galgodagedon\}@swansea.ac.uk}
}

\begin{document}
\maketitle
\begin{abstract}
Recent advances in language models have substantially improved Natural Language Understanding (NLU). Although widely used benchmarks suggest that Large Language Models (LLMs) can effectively disambiguate, their practical applicability in real-world narrative contexts remains underexplored. \textbf{SemEval-2026 Task 5} addresses this gap by introducing a task that predicts the human-perceived plausibility of a word sense within a short story. In this work, we propose an LLM-based framework for plausibility scoring of homonymous word senses in narrative texts using a structured reasoning mechanism. We examine the impact of fine-tuning low-parameter LLMs with diverse reasoning strategies, alongside dynamic few-shot prompting for large-parameter models, on accurate sense identification and plausibility estimation. Our results show that commercial large-parameter LLMs with dynamic few-shot prompting closely replicate human-like plausibility judgments. Furthermore, model ensembling slightly improves performance, better simulating the agreement patterns of five human annotators compared to single-model predictions.

\end{abstract}

\section{Introduction}

With the introduction of Transformer models \cite{NIPS2017_3f5ee243}, language models became increasingly capable at a variety of Natural Language Processing (NLP) tasks. Lexical ambiguity became a major challenge for traditional models due to the occurrence of two or more completely unrelated senses sharing the same spelling or pronunciation \cite{bevilacqua2021recent}. Even though recent studies reveal that large language models (LLMs) perform strongly at Word Sense Disambiguation (WSD) on common words and senses, their performance curtainls in identifying rare cases or less frequent words \cite{sumanathilaka2024assessing, meconi2025large}.

The existing benchmarks predominantly operate at the sentence level, where disambiguating homonyms are mostly based on neighbour-word clues, global context analysis, and syntactic clues and dependency relations \cite{raganato2017word, ballout2024fool, blevins-etal-2021-fews}. While these datasets are effective in constrained settings, this formulation has inherent limitations, such as isolated sentences often providing insufficient contextual evidence, failing to reflect the richer, multi-sentence contexts required for realistic real-world language understanding.

To address this gap, \textbf{SemEval-2026 Task 5} introduces the AmbiStory dataset \cite{gehring-roth-2025-ambistory}, which comprises narrative texts that naturally encode ambiguity through their discourse structure. Each instance consists of a short narrative with four to five sentences providing situational context (the precontext), followed by an ambiguous target sentence and, optionally, a concluding sentence. By modelling ambiguity within narrative settings, this benchmark enables a more realistic evaluation of the applicability of WSD algorithms in real-world scenarios. The source code for our implementation is publicly available at \url{https://github.com/Sumanathilaka/SwanNLP-at-SemEval-2026-Task-5}.

\subsection{Task Overview}

SemEval-2026 Task 5 \cite{semeval2026-task-5}, “Rating Plausibility of Word Senses in Ambiguous Sentences through Narrative Understanding”, is designed to evaluate the ability of computational models to simulate human reasoning when interpreting the sense of a homonym in a narrative context. Participants are required to predict a human-perceived plausibility score (1-5) for each story. Each instance consists of a precontext that grounds the narrative, an ambiguous sentence containing a homonym, and, optionally, an ending that often implies a particular word sense.

The underlying dataset, Ambistory, was annotated by human participants recruited via Prolific, with five independent annotations per instance on a five-point plausibility scale. Participants are provided with training, development, and test splits. The primary evaluation metrics are the Spearman correlation between predicted scores and the average human judgment and Accuracy within Standard Deviation, defined as the proportion of model predictions falling within at least one standard deviation of the annotators’ mean score. Dataset statistics are summarized in Table \ref{tab:Dataset Information}.

\begin{table}[t]
\centering
\caption{Dataset Statistics. Unique senses represent number of unique homonyms captured in each set.}
\label{tab:Dataset Information}
\begin{tabular}{lcc}
\hline
\textbf{Ambistory} & \textbf{\# Stories} & \textbf{\# Unique senses}\\
\hline

Train        & 2322 & 188 \\
Dev         & 600 & 42\\
Test       & 942 & 76\\
\hline
\end{tabular}
\end{table}

\section{Related Work}

Recent studies show that LLMs are substantially more effective at disambiguating commonly used homonyms, due to their ability to model rich contextual and semantic cues \cite{cahyawijaya2024thank, meconi2025large}. Although parameter-efficient models such as Qwen and Gemma lack the deep contextual representations of large-scale LLMs, fine-tuning has been shown to yield significant performance gains over their base versions \cite{basile2025exploring}, suggesting a more energy-efficient alternative for domain or task-specific disambiguation. These findings motivate our use of a supervised fine-tuning framework to simulate human plausibility judgments. In particular, we model both single-annotator and aggregated multi-annotator perspectives through a reasoning-driven pipeline that incorporates difficulty-aware analysis.

Prior work has also explored reformulating WSD as a higher-level reasoning task. \citet{sainz_what_2023} cast WSD as a textual entailment problem, prompting models to assess the compatibility between candidate sense descriptions and sentences containing ambiguous words. This zero-shot formulation outperforms random baselines and, in some cases, rivals supervised WSD systems. Complementarily, \citet{sumanathilaka_glossgpt_2025} investigates prompt engineering and in-context learning strategies with GPT-3.5-Turbo and GPT-4-Turbo can substantially improve the disambiguation task, while subsequent benchmarking identifies GPT-based models and DeepSeek as particularly effective for WSD \cite{sumanathilaka2024can}. Collectively, these studies indicate that LLMs are well-suited for sense disambiguation in context and that reasoning-oriented prompting can further enhance performance. Building on this line of work, our approach aims to enhance the challenging task of plausibility score prediction by leveraging dynamic few-shot extraction for in-context learning with larger models.

\section{Methodology}

In this study, we have evaluated three sets of approaches for predicting plausibility scores. We have used a supervised fine-tuning approach (SFT) with low-parameter LLMs, In-context learning via dynamic Retrieval Augmented Generation (RAG) with Chain-of-thought (CoT) reasoning \cite{wei2022chain} and model ensembling to further simulate the multi-annotator agreement. To support different reasoning processes across all approaches, we have classified the data into plausibility levels for use in each phase.

To implement the finetuning logic, we used average scores to identify the possible outcomes, which were determined by human annotators during the annotation process. This inferred rationale was then consistently incorporated into the fine-tuning procedure to simulate human judgment.

\begin{itemize}[leftmargin=*,nosep]
    \item $\text{Average} \geq 4.0$: Meaning strongly fits the context and the ending, indicating high plausibility.
    \item $3.0 \leq \text{Average} < 4.0$: Meaning reasonably fits the context and the ending, indicating moderate plausibility.
    \item $2.0 \leq \text{Average} < 3.0$: Meaning shows a weak connection to the context and the ending, indicating slight plausibility.
    \item $\text{Average} < 2.0$: Meaning does not fit the context or the ending and is therefore not plausible.
\end{itemize}

These plausibility bands were defined through a data-driven discretization of the averaged human ratings, informed by the score distribution and the annotation characteristics of the dataset \cite{snow-etal-2008-cheap}.

\begin{figure}
    \centering
    \includegraphics[width=0.45\textwidth]{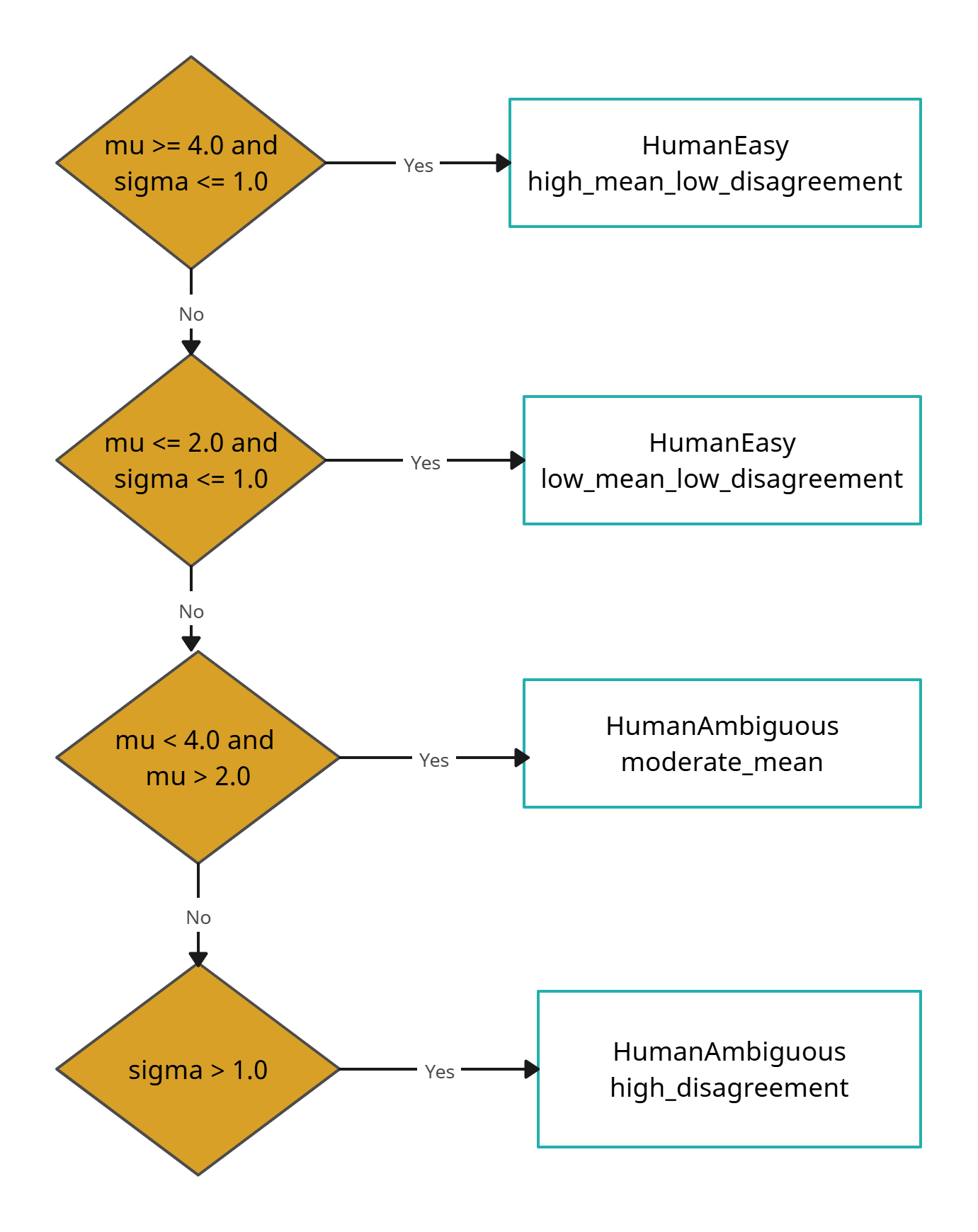}
    \caption{The flow used to classify whether a case is easy or difficult for humans.}
    \label{fig:image}
\end{figure}

\subsection{Finetuning low parameter models}

For the fine-tuning task, we employ four techniques to model the reasoning process. The fine-tuning procedure is incremental, incorporating multiple reasoning strategies proposed in prior work to improve WSD performance \cite{sumanathilaka2026exploration}. We adapt these approaches to better align with the task of simulating human annotation of plausibility scores. Each technique is specifically designed to reflect how humans evaluate and assign plausibility. The design strategies used in this study are outlined as follows.

\begin{itemize}[leftmargin=*,nosep]
\item Given a short story, a target ambiguous word, a candidate sense, and a narrative ending, the model assigns a single plausibility score corresponding to the judgment of one annotator.

\item Given a short story, a target ambiguous word, a candidate sense, and a narrative ending, the model generates five plausibility scores to simulate the judgments of five different annotators and capture variation in human reasoning.

\item Given a story context, a target ambiguous word, a candidate sense, and a narrative ending, the model first determines whether the candidate sense is plausible by analyzing the surrounding contextual cues and the narrative ending. It then assigns the most appropriate plausibility score based on this reasoning.

\item Given a story context, a target ambiguous word, a candidate sense, and a narrative ending, the model first identifies the difficulty level of the instance, such as human-easy or human-ambiguous, by analyzing the characteristics of the story. It then reasons about the candidate sense and assigns a plausibility score accordingly.
\end{itemize}

The models were fine-tuned to simulate both single-annotator and multi-annotator scoring processes in line with the design goals described above. The fine-tuning inputs, including the reasoning rationale and the expected output format for each strategy, are presented in Table \ref{tab:finetuning_strategies}. In the third fine-tuning strategy, we used GLOSSGPT \cite{sumanathilaka_glossgpt_2025}, a state-of-the-art word sense disambiguation system, to provide a likely sense interpretation for each homonymous target word based on the story context. This sense prediction was precomputed and used as an auxiliary reasoning cue during fine-tuning, rather than treated as a definitive ground-truth label.

More specifically, GLOSSGPT was used to identify the most probable sense supported by the contextual clues in the story, which helped structure the intermediate reasoning process for plausibility scoring. However, because plausibility judgments may vary across annotators, especially in human-ambiguous cases, GLOSSGPT's role was limited to providing a stable semantic reference point. The fine-tuning rationale was designed to reduce over-reliance on a single predicted sense and instead emphasize how low-parameter LLMs can model human-like variation in plausibility assessment, particularly when the context supports more than one reasonable interpretation.

\begin{table*}[t]
\centering
\caption{Mapping of fine-tuning strategies to their inputs for training and expected outputs during inference.}
\small
\begin{tabular}{p{4.2cm} p{5.2cm} p{5.2cm}}
\hline
\textbf{Approach} & \textbf{Input} & \textbf{Expected Output} \\
\hline

Finetuning for Single Annotator Simulation &
Story context, target ambiguous word, candidate sense, and narrative ending. &
A plausibility score reflecting the decision of a single human annotator. \\
\hline
Finetuning for Five Annotator Simulation &
Story context, target ambiguous word, candidate sense, and narrative ending. &
An aggregated plausibility score simulating the consensus of five annotators. \\
\hline
Finetuning for Single Annotator with Thinking &
Story context, target ambiguous word, candidate sense, narrative ending , Correct sense and Reasoning steps.&
A plausibility score guided by explicit reasoning steps for a single annotator. \\
\hline
Finetuning for Single Annotator with Difficulty Analysis &
Story context, target ambiguous word, candidate sense, and narrative ending, with difficulty tag precomputed using the approach presented in the figure \ref{fig:image}.&
A plausibility score for a single annotator. \\

\hline
\end{tabular}
\label{tab:finetuning_strategies}
\end{table*}

\subsubsection{Study Setup}

For model development, supervised fine-tuning (SFT) was employed. The baseline models were fine-tuned using Low-Rank Adaptation (LoRA) to enable efficient training while reducing computational overhead. Training data were pre-processed into a desired chat-style prompt–response format, with each query containing homonym, pre-context, sentence and ending. For the last two phases of the finetuning, related pre-computed senses and difficulty tags were incorporated to build the reasoning logic.

Fine-tuning was conducted using Hugging Face transformers and trl libraries, with a custom prompt formatting function to standardize inputs. The data were tokenized using the native tokenizer of each model. We used a batch size of 
4, gradient accumulation over 8 steps, and a learning rate of $=$ $2 \times 10^{-4}$  with these settings chosen based on preliminary experimentation. Optimization was performed using AdamW with a linear learning-rate scheduler. All experiments were conducted on an NVIDIA A100-PCIE-40GB GPU, which provided sufficient memory to fine-tune the models efficiently without quantization in the final configuration.

\subsection{Dynamic Few-shot learning for Commercial Models}

During the first phase of the study, we observed that small models, including fine-tuned variants, are not well suited to handling ambiguous sentences that humans judge as moderately plausible. To address this limitation, we propose a RAG-inspired approach \cite{lewis2020retrieval} that enriches in-context knowledge during inference in large-parameter models. Following the logic illustrated in Figure~\ref{fig:image}, we subdivided the training data into three categories.

\textbf{Ambiguous Context (1088 Records):}
This category contains instances for which human annotators assigned diverse plausibility scores. Using the annotations' mean and standard deviation, we identified cases with substantial disagreement among annotators, indicating high ambiguity.

\textbf{Human Easy - High Score (631 Records):}
This category consists of instances that humans can annotate with high confidence. When all annotators are in agreement and consistently assign high scores, it indicates that the given meaning is highly suitable for the ambiguous sentence.

\textbf{Human Easy - Low Score (561 Records):} 
This category also includes instances that are easy for humans to annotate, but where annotators consistently assign low scores. Such an agreement indicates that the given meaning is not suitable for the homonym in the provided context.

This categorization formed the basis for constructing the vector stores used as retrievers in our few-shot inference setting. Each story was treated as a separate chunk in the vector store, and embeddings were generated using \textit{BAAI/bge-small-en-v1.5}. A FAISS vector index was used to store the story embeddings. The retriever employs \textit{similarity\_search} to fetch the most relevant few-shot examples during the inference process.

Initial experiments were conducted with K=1,2,3, and we observed that incorporating additional contextual examples slightly improves in-context learning accuracy. However, considering the computational cost of inference, we limited our experiments to K=1 and 3. All the experiments were conducted keeping the temperature at 0 to ensure the replicability of the study \cite{sumanathilaka2025exploring}.

For the baseline evaluation, we intentionally omitted retrieval steps and the few-shot examples and followed the original instructions and scoring rubric exactly as defined for the few-shot experiments. No additional demonstrations or auxiliary guidance were introduced in the baseline prompts. This design choice ensured a fair comparison between the baseline and the proposed dynamic few-shot setting, in which improvements could be attributed to the inclusion of retrieved in-context examples rather than to changes in the underlying task instructions. The detailed prompt used in the inference can be found in the Appendix \ref{sec:appendix A}.

\subsection{Model ensembling to simulate multi-annotator agreement}

The previous phase of our experiments shows that Human Easy cases, including both high- and low-plausibility scores, are relatively easy to simulate, whereas ambiguous context cases struggle to accurately reproduce the plausibility scores assigned by five human annotators. To address this limitation, we propose an ensemble-based approach in which each model used in phases 1 and 2 acts as an individual annotator. We explored multiple strategies to simulate the five-annotator process in this phase and achieved a considerable improvement in the final performance. Since the training data had already been used for fine-tuning and vector database construction, we restricted the training of the ensemble models to the development set. To ensure robust evaluation under this constraint, we employed k-fold cross-validation on the development data and final results were reported as the average performance across all folds. 

For model ensembling, we employed the following approaches:

\paragraph{Majority Voting:} The final plausibility score was determined by selecting the most frequently predicted score across the five models. When multiple scores received the same maximum number of votes, we computed the mean of the tied scores and rounded it to the nearest integer.

\paragraph{Equal-Weight Averaging:} All five model scores were averaged with equal weights (0.2), and the resulting value was rounded to the nearest integer.

\paragraph{Performance-Based Weighted Averaging:} Model weights were assigned based on their relative performance in Phases 1 and 2 on the development set. The final prediction was obtained as a weighted average of the individual model outputs, then rounded to the nearest integer.

\paragraph{Linear Regression Ensemble (LRE):} A linear regression model was trained using the predictions of the five base models as input features and the gold plausibility scores as targets. The learned regression coefficients implicitly capture the relative importance of each model.

\paragraph{Support Vector Regression (SVR):} We employed a Support Vector Regression–based meta-ensemble with an RBF kernel. Since SVR is sensitive to feature scaling, model predictions were standardized prior to training. To ensure robustness, performance was assessed using 5-fold cross-validation on the development set.

\paragraph{XGBoost Ensemble:} An XGBoost regressor was trained as a meta-learner over the phase 1 and 2 predictions. Optimal hyperparameters were selected via grid search on the development set. The final model was evaluated exclusively on the test data.

\section{Results and Discussion}

\textbf{Performance of Finetuned Low-parameter LLMs} is summarized in Table~\ref{tab:matrix_results_smaller}. The strongest results were achieved by difficulty-analysis-based reasoning models, indicating that explicitly identifying the difficulty of a story and its ending helps better simulate human judgment when assigning plausibility scores. Compared to general score prediction based on fine-tuning, reasoning-driven strategies show relatively higher performance. Importantly, Qwen-4B outperforms Gemma-4B in all experiments. Simulations based on five annotators consistently underperform compared to single-annotator simulations. The qualitative analysis further suggests that, even for content perceived as easy by humans, models tend to introduce additional diversity to mimic human behavior, which can negatively impact overall performance. In contrast, teaching smaller LLMs to reason using a structured CoT approach leads to performance improvements, highlighting a promising direction for future research.

\begin{table*}[t]
\centering
\caption{Evaluation results of Gemma 4B and Qwen 4B across different fine-tuning and prompting strategies. Sc denotes Spearman correlation. Acc denotes accuracy. Best results for each approach are highlighted in bold.}
\small
\begin{tabular}{l l c c | c c c c}
\hline
\textbf{Model} & \textbf{Split} & \textbf{Metric} &
\textbf{Base} &
\multicolumn{4}{c}{\textbf{Fine-tuned Simulation}} \\
& & &  \textbf{inference} &
\textbf{Single Annotator} &
\textbf{Five Annotator} &
\textbf{Reasoning} &
\textbf{Difficulty Analysis} \\
\hline

\multirow{4}{*}{Gemma 4B}
& \multirow{2}{*}{Dev}
& Sc
& 0.034 & 0.279 & 0.127 & 0.373 & \textbf{0.467} \\
& 
& Acc
& 0.424 & 0.559 & 0.452 & 0.605 & \textbf{0.621} \\

& \multirow{2}{*}{Test}
& Sc
& 0.232 & 0.316 & 0.202 & 0.314 & \textbf{0.424 }\\
&
& Acc
& 0.519 & 0.565 & 0.494 & 0.575 & \textbf{0.623} \\
\hline

\multirow{4}{*}{Qwen 4B}
& \multirow{2}{*}{Dev}
& Sc
& 0.273 & 0.265 & 0.316 & 0.281 & \textbf{0.500} \\
&
& Acc
& 0.573 & 0.565 & 0.498 & 0.575 & \textbf{0.667} \\

& \multirow{2}{*}{Test}
& Sc
& 0.031 & 0.242 & 0.369 & 0.376 & \textbf{0.491} \\
&
& Acc
& 0.414 & 0.550 & 0.528 & 0.586 & \textbf{0.631 }\\
\hline
\end{tabular}
\label{tab:matrix_results_smaller}
\end{table*}

\begin{table*}[t]
\centering
\caption{Evaluation results of Deepseek V3, GPT 4o, and Gemini 2.5 flash-lite across one-shot and three-shot at each category. Sc denotes Spearman correlation. Acc denotes accuracy. Best results for the approach are highlighted in bold.}
\small
\begin{tabular}{l l c c | c c }
\hline
\textbf{Model} & \textbf{Split} & \textbf{Metric} &
\textbf{Zero} &
\multicolumn{2}{c}{\textbf{Dynamic few shot learning}} \\
& & &  \textbf{shot} &
\textbf{K=1} &
\textbf{K=3} \\
\hline

\multirow{4}{*}{Deepseek v3}
& \multirow{2}{*}{Dev}
& Sc
& 0.676 & 0.674 & 0.685  \\
& 
& Acc
& 0.792 & \textbf{0.809} & \textbf{0.821 }\\

& \multirow{2}{*}{Test}
& Sc
& 0.642 & 0.641 & 0.674 \\
&
& Acc
& 0.667 & 0.708 & 0.708 \\
\hline

\multirow{4}{*}{GPT 4o}
& \multirow{2}{*}{Dev}
& Sc
& \textbf{0.743} & \textbf{0.724} & \textbf{0.726} \\
&
& Acc
&\textbf{0.804} & 0.803 & 0.812 \\

& \multirow{2}{*}{Test}
& Sc
& \textbf{0.741} & \textbf{0.746 }&  \textbf{0.755} \\
&
& Acc
& \textbf{0.792} & \textbf{0.802} &  \textbf{0.798} \\
\hline

\multirow{4}{*}{Gemini 2.5 flash-lite}
& \multirow{2}{*}{Dev}
& Sc
& 0.664 & 0.672 & 0.682  \\
&
& Acc
& 0.721 & 0.736 & 0.750  \\

& \multirow{2}{*}{Test}
& Sc
& 0.674 & 0.706 & 0.531  \\
&
& Acc
& 0.647 & 0.675 & 0.636 \\
\hline

\end{tabular}
\label{tab:matrix_results_larger}
\end{table*}

\begin{table*}[t!]
\centering
\caption{Evaluation results of the Ensemble approach. Sc denotes Spearman correlation. Acc denotes accuracy. Best results per model are highlighted in bold.}
\small
\begin{tabular}{l c c c c }
\hline
\textbf{Ensemble approach} & \multicolumn{2}{c}{\textbf{Dev}}  & \multicolumn{2}{c}{\textbf{Test}} \\
& Sc & Acc & Sc & Acc \\
\hline
Majority vote & 0.687 & 0.738 & 0.713 & 0.779 \\
Similar weight analysis & 0.741 & 0.799 & 0.701  & 0.787  \\
Weights by relative performance  &  0.737 & 0.838 & 0.701 & 0.768 \\
Linear Regression Ensemble   & 0.769 & 0.841 & 0.723 &\textbf{ 0.797} \\
Support Vector Regression  & \textbf{0.821} & \textbf{0.879} & 0.713 & 0.771 \\
XGBOOST   & - & - & \textbf{ 0.724 } & 0.780 \\

\hline
\end{tabular}
\label{tab:matrix_results_ensemble}
\end{table*}


\paragraph{Performance of Dynamic Few-shot Learning} demonstrates clear improvements over the fine-tuning of smaller models. Among all evaluated systems, GPT-4o consistently outperforms the others in both SC and accuracy across most settings, highlighting its strong capability to approximate human judgment in plausibility scoring tasks. Compared to zero-shot inference, dynamic few-shot prompting using the proposed approach yields consistent, though modest, gains on both the development and test sets. This suggests that incorporating in-context examples effectively enhances the model’s ability to simulate plausibility judgments.

Interestingly, the optimal number of shots varies across splits: one-shot prompting (with one example per category) achieves the best performance on the test set, whereas three-shot prompting performs best on the development set. This indicates a trade-off between generalization and overfitting to in-context examples, where increasing the number of shots may benefit validation performance but does not always translate to improved test-time robustness. 


\paragraph{Performance of the Ensembling approach} was evaluated in the final phase of the study to simulate the five-annotator agreement process using the five models. While the Linear Regression Ensemble achieves the best overall accuracy on the test set (0.797), the best Spearman score was achieved by XGBOOST under the optimal parameters selected via grid search (0.724), slightly outperforming LRE. The SVR model performs best on the development set, indicating the presence of non-linear relationships in the data. Overall, the ensembling results demonstrate consistent performance improvements across experiments, indicating that multi-annotator agreement-based simulation can yield better outcomes than single-annotator simulation.

\section{Conclusion}

We achieved an overall \textbf{10th} place on the leaderboard, resulting in an average score of \textit{0.760}. These results demonstrate the effectiveness of our approach in modelling plausibility within the narrative WSD task. Our study shows that plausibility scoring is best captured through reasoning-aware approaches that explicitly simulate human decision-making processes. In particular, LLMs that leverage dynamic few-shot prompting closely approximate human plausibility judgments, while structured reasoning strategies and ensemble methods further enhance robustness and performance. 
Despite these advances, our results also highlight important limitations in current approaches. While the models exhibit a reasonable capacity to approximate human reasoning, their performance declines in ambiguous or context-sensitive scenarios where plausibility judgments are inherently uncertain. This suggests that, although LLMs are increasingly effective at modelling human-like reasoning, they still struggle with consistency and fine-grained interpretive nuance, leaving considerable room for improvement in achieving more reliable and human-aligned judgments.

\section*{Acknowledgments}

We acknowledge the support of the Supercomputing Wales project, which is part-funded by the European Regional Development Fund (ERDF) via the Welsh Government. 

\bibliography{custom}

\appendix
\section{Appendix}
\label{sec:appendix A}

\begin{table}[hbt!]
\centering
\caption{Prompt used for plausibility scoring with Larger models.}
\small
\begin{tabular}{p{0.99\linewidth}}
\hline
\\
Your task is to rate the plausibility of a word's meaning on a scale of 1--5 
based on a short story. You must follow the Thinking Process below to arrive at your score. You will be provided with few example stories that illustrate the scoring rubric for different levels of plausibility, based on similar stories: \{Few-shot\_Examples\}.
\\ \\
Now evaluate the following story and proposed meaning:

\textbf{Precontext:} \{precontext\} \\
\textbf{Sentence: }\{sentence\} \\
\textbf{}\textbf{Ending:} \{ending\}

\textbf{Word:} ``\{homonym\}'' \\
\textbf{Proposed Meaning to Evaluate: }``\{judged\_meaning\}''

\\ \\
\textit{Complete each step of this process in your analysis.}

\textbf{Instructions:}
\begin{enumerate}[leftmargin=*,nosep]
\item  Analyze the Context: Read the complete story and identify all clues that might support or contradict the 'Proposed Meaning'.
\item  List Evidence For:State the parts of the story that make the 'Proposed Meaning' plausible.
\item List Evidence Against: State any parts of the story that make the 'Proposed Meaning' implausible.
\item  Synthesize and Score: Based on the evidence, provide a final plausibility score using the rubric below.
\end{enumerate}
\\ \\
\textbf{Scoring Rubric:}
\begin{itemize}[leftmargin=*,nosep]
    \item \textbf{5: Perfectly plausible.} The meaning is strongly supported by the entire context, and all parts of the story form a consistent, logical narrative.
    \item \textbf{4: Very plausible.} The meaning fits well and is consistent. There might be minor ambiguity, but no real contradictions.
    \item \textbf{3: Moderately plausible.} The meaning is possible, but the context is ambiguous or contains minor conflicting clues.
    \item \textbf{2: Barely plausible.} The meaning largely conflicts with the context.
    \item \textbf{1: Implausible.} The meaning is directly and strongly contradicted by the context.
\end{itemize}
\\ \\
Do all the reasoning mentally / privately --- do not print it; print only the final integer score as an output.
\\
\hline
\end{tabular}
\label{tab:prompt}
\end{table}

\end{document}